\documentclass[10pt,twocolumn,letterpaper]{article}
\usepackage{wacv}
\usepackage{times}
\usepackage{epsfig}
\usepackage{graphicx}
\usepackage{amsmath}
\usepackage{amssymb}
\usepackage{tikz}

\usepackage{multirow}
\usepackage[utf8]{inputenc}
\usepackage{xspace}
\usepackage[english]{babel}
\usepackage{subfig}
\usepackage{tabularx}
\usepackage{array}
\newcolumntype{L}[1]{>{\raggedright\let\newline\\\arraybackslash\hspace{0pt}}m{#1}}
\newcolumntype{C}[1]{>{\centering\let\newline\\\arraybackslash\hspace{0pt}}m{#1}}
\newcolumntype{R}[1]{>{\raggedleft\let\newline\\\arraybackslash\hspace{0pt}}m{#1}}

\newcommand{\imgI}{\mathcal{I}}
\newcommand{\imgJ}{\mathcal{J}}
\newcommand{\mat}[1]{\mathbf{#1}}
\newcommand{\definedas}{\stackrel{def}{=}}

\wacvfinalcopy 
\def\wacvPaperID{21} 

\ifwacvfinal\fi
\begin{document}

\title{Joint Epipolar Tracking (JET):\\ Simultaneous optimization of epipolar geometry and feature correspondences}
\author{Henry Bradler\textsuperscript{1}, Matthias Ochs\textsuperscript{1}, Nolang Fanani\textsuperscript{1}\\
\textsuperscript{1}Visual Sensorics \& Information Processing Lab\\
Goethe University, Frankfurt, Germany\\
{\tt\small \{bradler,ochs,fanani,mester\}@vsi.cs.uni-frankfurt.de}
\and
Rudolf Mester\textsuperscript{1,2}\\
\textsuperscript{2}Computer Vision Laboratory, ISY\\
Link\"oping University, Sweden\\
{\tt\small mester@isy.liu.se}
}
\maketitle

\begin{tikzpicture}[remember picture,overlay]
\node[anchor=south,yshift=10pt] at (current page.south) {\fbox{\parbox{\dimexpr\textwidth-\fboxsep-\fboxrule\relax}{\footnotesize \textcopyright 2017 IEEE. Personal use of this material is permitted.
			Permission from IEEE must be obtained for all other uses, in any current or future
			media, including reprinting/republishing this material for advertising or promotional
			purposes, creating new collective works, for resale or redistribution to servers or
			lists, or reuse of any copyrighted component of this work in other works.
			DOI: pending}}};
\end{tikzpicture}

\ifwacvfinal\thispagestyle{empty}\fi

\begin{abstract}

Traditionally, pose estimation is considered as a two step problem. First, feature correspondences are determined by direct comparison of image patches, or by associating feature descriptors. In a second step, the relative pose and the coordinates of corresponding points are estimated, most often by minimizing the reprojection error (RPE). RPE optimization is based on a loss function that is merely aware of the feature pixel positions but not of the underlying image intensities. In this paper, we propose a sparse direct method which introduces a loss function that allows to simultaneously optimize the unscaled relative pose, as well as the set of feature correspondences directly considering the image intensity values. Furthermore, we show how to integrate statistical prior information on the motion into the optimization process. This constructive inclusion of a Bayesian bias term is particularly efficient in application cases with a strongly predictable (short term) dynamic, \eg in a driving scenario. In our experiments, we demonstrate that the `JET` algorithm we propose outperforms the classical reprojection error optimization on two synthetic datasets and on the KITTI dataset. The JET algorithm runs in real-time on a single CPU thread.

\end{abstract}

\section{Introduction}

\begin{figure}[h!]
	\begin{center}
		\includegraphics[width=1.0\linewidth]{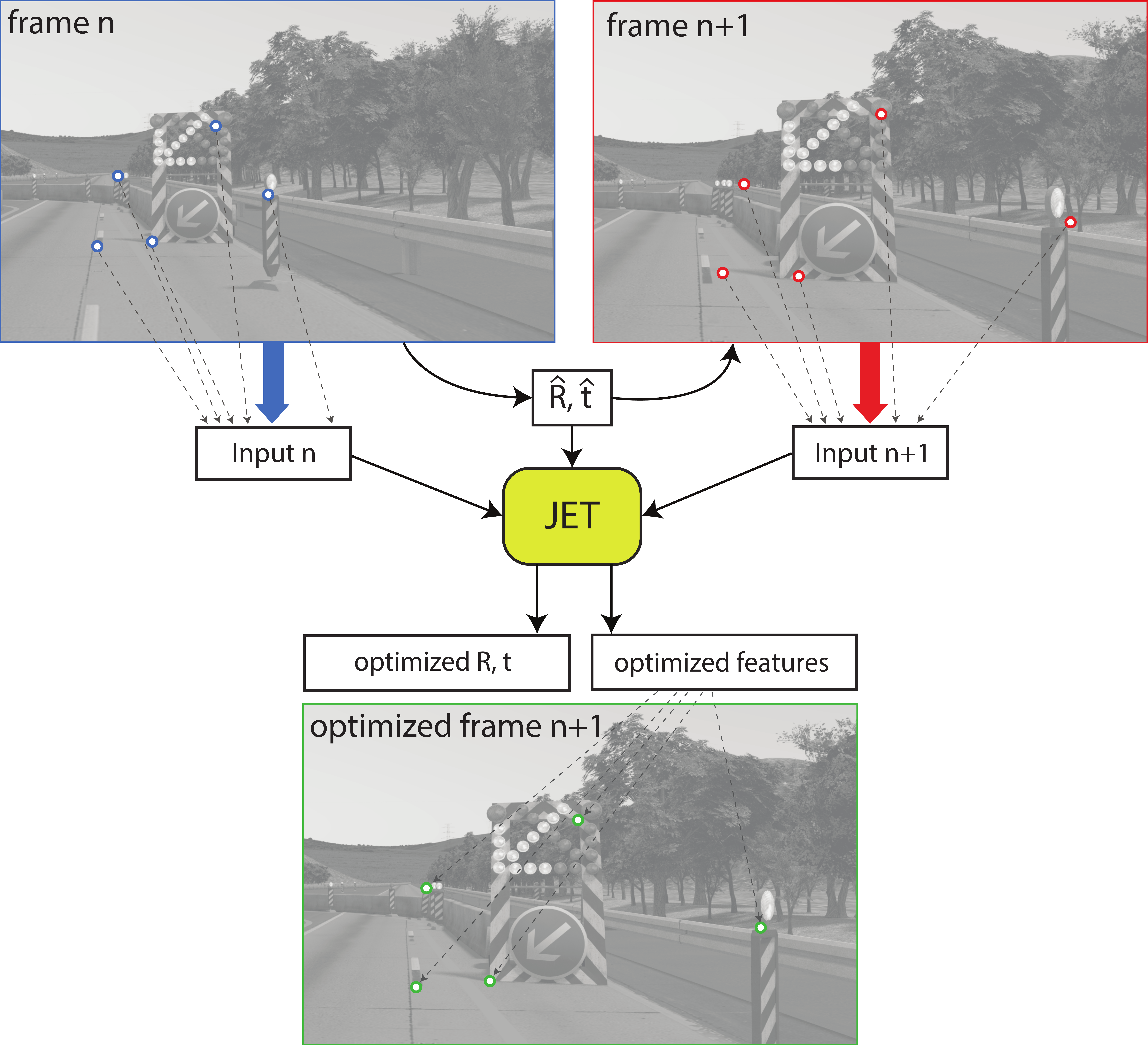}
	\end{center}
	\caption{Flow chart of the JET algorithm.}
\label{fig:jet-overview}	
\end{figure}
The main contribution of this work is the introduction of a joint loss function which is based on the photometric error of all feature correspondences. The correspondences are parameterized by one underlying epipolar geometry. This guarantees all correspondences to be epipolar-conform by construction, and allows to directly optimize the pose based on image intensities. Starting point is the well known Lucas-Kanade tracking method \cite{LucasKanade1981IJCAI} which employs a quadratic photometric loss function (SSD) on a \emph{single} image patch to optimize feature correspondences. Given the epipolar geometry, the search space can be drastically decreased from a 2D to a 1D search by including an epipolar constraint \cite{PicciniPerssonNordbergFelsbergMester2014ECCV,TrummerDenzler2008VISAPP}.

We show how to optimize all correspondences simultaneously \emph{and} optimize the epipolar geometry at the same time, given coarse initial values of these entities (the typical situation in many applications). This is achieved by varying the epipolar geometry, and by this the adjusted correspondences of all features, in a way that minimizes the \emph{joint} photometric loss function. We denote the resulting procedure as \emph{Joint Epipolar Tracking} (JET). As this joint optimization is performed directly on the intensity values, JET is a  \emph{`direct`} method, in today's terminology. This is in contrast to the widely used minimization of the reprojection error which
distills photometric information into geometric information and subsequently disregards the mere image intensities.

We show that JET outperforms the standard minimization of the reprojection error (RPE), when optimizing the relative pose (ego motion of the camera). The comparison was performed on synthetic and real data sets (all publicly available), synthetic data sets in order to have perfect ground truth and real data sets to demonstrate the feasibility under realistic conditions. The synthetic data sets are COnGRATS \cite{BiedermannOchsMester2015IVCNZ} (driving scenes on a road scene) and RGB-D data from ICL-NUIM \cite{HandaWhelanMcDonaldDavison2014ICRA} (indoor footage of a hand-held camera). As a representative of real data, we utilize the well known KITTI dataset \cite{GeigerLenzStillerUrtasun2013IJRR,GeigerLenzUrtasun2012CVPR}, which consists of real driving scenes in urban and highway scenarios. As dense depth information or optical flow ground truth are not available for this data set, we focus on comparing the quality of JET against RPE by regarding the relative pose.

Since we regard monocular image data, the scale of the pose remains undetermined and we only analyze the relative rotation and the relative \emph{unscaled} translation of the motion. For a calibrated camera, these entities will entirely define the epipolar geometry as it is scale independent as well.
\section{Related Work}

Approaches to relative pose or 3D motion estimation can be divided into two basic categories: \emph{feature-based} methods and \emph{direct} methods, with some hybrid approaches existing as well. Feature-based methods are characterized by the extraction and the matching of salient and reproducible features that are tracked over frames. Prominent examples of feature point based optimization methods are \cite{BadinoYamamotoKanade2013}, \cite{PerssonPicciniFelsbergMester2015}, and \cite{CvisicPetrovic2015}. Usually, these approaches minimize the reprojection error of tracked feature points.

So called `direct`, appearance- or intensity-based methods, on the other hand, operate directly by matching pixel intensities. They propagate the original image information into the optimization scheme, usually using a differential optimization approach and therefore can often provide more accurate estimates of pose and structure. DTAM \cite{NewcombeLovegroveDavison2011} was among the first real time dense systems. Semi-Dense Visual Odometry for a Monocular Camera \cite{EngelSturmCremers2013} and its successor LSD-SLAM \cite{EngelSchoepsCremers2014} as well as SVO \cite{ForsterPizzoliScaramuzza2014} are more recent examples. We share the opinion of the authors of \cite{EngelSturmCremers2013} who state that the separation into feature detection and tracking versus a state estimation creates an artificial gap between the data and the state sought. 

PTAM \cite{KleinMurray2007} could also be considered as a hybrid method: A weak motion prior is used to initialize the search for a small number of features using a modified KLT at the highest level of a scale pyramid. The resulting tracks provide a better egomotion prediction which is then used to search for a larger number of KLT tracks at lower pyramid levels and so on until the bottom level is reached. 

Stabilizing the estimation of correspondences by integrating a given (or assumed) epipolar relation into the matching process has been used in numerous approaches. For instance, the authors of \cite{AlismailBrowningLucey2016,PicciniPerssonNordbergFelsbergMester2014ECCV,TrummerDenzlerMunkelt2009SCIA,VogelSchindlerRoth2011,YamaguchiMcAllesterUrtasun2013}
use epipolar constraints for stabilizing discrete matching, whereas Valgaerts et al. \cite{ValgaertsBruhnWeickert2008} proposed a variational approach to estimate dense optical flow and the epipolar geometry (represented by the fundamental matrix) simultaneously. Other direct methods that also explicitly take into consideration the depth structure of the scene are \cite{BergerNeufeldBeckerLenzenSchnoerr2015SSVM,LenzenBerger2015SSVM,NeufeldBergerLenzenSchnoerr2015GCPR}.

An important property which typically distinguishes appearance, direct, dense or semi-dense from feature based approaches is that direct methods often use parametric models of the flow field  and hence can utilize edges as well as corners. If no explicit motion priors or dynamic models are used, these direct methods generally depend on a high frame rate that ensures moderate displacements, whereas feature-based matching can work even with very large displacements. However, even in this case a photometric `direct` post-optimization can be performed. JET is a well suited method to do just this.
\section{Approach}

We outline our approach and introduce our notation, starting from plain Lucas-Kanade tracking \cite{LucasKanade1981IJCAI} in section \ref{sec:general-lucas-kanade-tracking} and subsequently revisiting epipolar constrained KL tracking in section \ref{sec:epipolar-constrained-tracking}. This leads to the presentation of joint epipolar tracking in section \ref{sec:joint-epipolar-tracking}.

\subsection{General Lucas-Kanade Tracking}
\label{sec:general-lucas-kanade-tracking}

The aim of differential direct tracking, often denoted as Lucas-Kanade tracking \cite{LucasKanade1981IJCAI} is to successively determine the corresponding image feature point position $\vec{y}_{k} = \vec{x}_{k} + \vec{v}_{k}$ in image $\imgJ$ for a given feature point $\vec{x}_{k}$ in another image $\imgI$.
We use the weighted sum of squared differences (WSSD) as loss function
for patch comparisons, thus implicitly modeling the image noise as
signal-independent, i.i.d. and Gaussian.
\begin{align}
	Q_{k} \definedas \sum_{\vec{x}} \mat{W}[\vec{x} - \vec{x}_{k}] \cdot \left( \imgI[\vec{x}] - \imgJ[\underbrace{\vec{x} + \vec{v}_{k}^{0}}_{\vec{y}}] \right)^{2}.
    \label{eq:weighted-ssd}
\end{align}
The non-negative pixel weights and the size of the patches are defined by a normalized kernel $\mat{W}$. All points $\vec{x}$ and $\vec{y} = \vec{x} + \vec{v}_{k}^{0}$ with a non-zero weight $\mat{W}[\vec{x} - \vec{x}_{k}]$ are taken into account for the patch difference. In a typical scenario, a feature point $\vec{x}_{k}$ and an initial estimate $\vec{y}_{k} = \vec{x}_{k} + \vec{v}_{k}^{0}$ for the corresponding feature are given and the task is to optimize this correspondence by minimizing the WSSD
\begin{align}
    Q_{k}(\vec{v}_{k}) \stackrel{def}{=} \sum_{\vec{x}} \mat{W}[\vec{x} - \vec{x}_{k}] \cdot \left( \imgI[\vec{x}] - \imgJ[\vec{y} + \vec{v}_{k}] \right)^{2} \rightarrow \min
    \label{eq:exact-wssd}
\end{align}
for a specific realization of the image displacement $\vec{v}_{k}$. Since this problem cannot be solved directly in closed form, a local first order Taylor approximation of the image difference $\imgI[\vec{x}] - \imgJ[\vec{y} + \vec{v}_{k}]$ is usually applied. This yields the approximated weighted sum of squared differences: 
\begin{align}
	\widetilde{Q}_{k}(\vec{v}_{k}) &\stackrel{def}{=} \sum_{\vec{x}} \mat{W}[\vec{x} - \vec{x}_{k}] \cdot \left( \imgI[\vec{x}] - \imgJ[\vec{y}] - \vec{v}_{k}^{T} \cdot \frac{\partial\imgJ}{\partial\vec{x}}[\vec{y}] \right)^{2}
	\nonumber
	\\
	&= \vec{v}_{k}^{T} \cdot \mat{A}_{k} \cdot \vec{v}_{k} + 2 \cdot \vec{v}_{k}^{T} \cdot \vec{b}_{k} + c_{k},
	\label{eq:approx-wssd}
\end{align}
using the abbreviations
\begin{align}
	\mat{A}_{k} &\definedas \sum_{\vec{x}} \mat{W}[\vec{x} - \vec{x}_{k}] \cdot \left( \frac{\partial\imgJ}{\partial\vec{x}}[\vec{y}] \right) \cdot \left( \frac{\partial\imgJ}{\partial\vec{x}}[\vec{y}] \right)^{T},
	\nonumber
	\\
	\vec{b}_{k} &\definedas \sum_{\vec{x}} -\mat{W}[\vec{x} - \vec{x}_{k}] \cdot \frac{\partial \imgJ}{\partial \vec{x}}[\vec{y}] \cdot \left( \imgI[\vec{x}] - \imgJ[\vec{y}] \right),
	\nonumber
	\\
	c_{k} &\definedas \sum_{\vec{x}} \mat{W}[\vec{x} - \vec{x}_{k}] \left( \imgI[\vec{x}] - \imgJ[\vec{y}] \right)^{2}.
\end{align}
Since the image difference has been linearized, this is an approximation to the `true` optimization problem, well known in nonlinear optimization theory as the \emph{Gauss-Newton} method. The approximation in equation \eqref{eq:approx-wssd} yields a convex parabolic function which allows to solve for the optimal displacement $\vec{v}_{k}$. Due to the linearization of the image, this approximation should only be used to improve the feature correspondence which then serves as a new initialization for another step of the incremental optimization process.

\subsection{Epipolar Constrained Tracking}
\label{sec:epipolar-constrained-tracking}

In \emph{constrained epipolar tracking}, we consider the relative pose given by a rotation matrix $\mat{R}$ and a translation vector $\vec{t}$ to be known, and adjust the feature correspondences $\{\vec{x}_{k} \leftrightarrow \vec{y}_{k} + \vec{v}_{k}\}$ to comply with the given epipolar geometry. This yields the epipolar constraint:
\begin{align}
	&& (\vec{y}_{k}^{T} + \vec{v}_{k}^{T}, 1) \cdot \mat{F} \cdot 
	\begin{pmatrix}
		\vec{x}_{k}	\\
		1
	\end{pmatrix}
	&\stackrel{!}{=} 0
	\label{eq:epipolar-constraint}
	\\
	&\Leftrightarrow& \vec{v}_{k}^{T} \cdot \underbrace{
	\begin{pmatrix}
		1	&	0	&	0	\\
		0	&	1	&	0
	\end{pmatrix}
	\cdot \mat{F}}_{\stackrel{def}{=} \mat{F}^{'}} \cdot 
	\begin{pmatrix}
		\vec{x}_{k}	\\
		1
	\end{pmatrix}
	&\stackrel{!}{=} - (\vec{y}_{k}^{T}, 1) \cdot \mat{F} \cdot 
	\begin{pmatrix}
		\vec{x}_{k}	\\
		1
	\end{pmatrix}.
	\nonumber
\end{align}
In equation \eqref{eq:epipolar-constraint}, $\mat{F}$ is the fundamental matrix,
\begin{align}
	\mat{F} \definedas \mat{K}^{-T} \cdot \left[ \vec{t} \right]_{\times} \cdot \mat{R} \cdot \mat{K}^{-1},
	\label{eq:fundamental-matrix-definition}
\end{align}
which is defined up to a scale factor. $\mat{K}$ is the camera matrix holding the intrinsic camera parameters and $\left[ \vec{t} \right]_{\times}$ is the skew symmetric matrix of the translation vector. We write $\mat{F} = \mat{F}(\vec{p})$ to denote that, for a given camera matrix, the fundamental matrix is fully determined by the (unscaled) motion parameters $\vec{p}$. We use the polar parametrization of the rigid transformation as proposed in \cite{BradlerWiegandMester2015ICCV-CVRSUAD}:
\begin{align}
	\vec{p} \definedas (\theta, \psi, \phi, \alpha, \beta)^{T}.
	\label{eq:motion-parameters}
\end{align}
These parameters are a minimum representation of the relative unscaled pose. The pitch angle $\theta$, the yaw angle $\psi$ and the roll $\phi$ are the rotational degrees of freedom about the $x$-, $y$- and $z$-axis. The azimuth $\alpha$ and the polar angle $\beta$ represent the unscaled translation $\vec{t}$ in polar coordinates. 

Using Lagrange multipliers, we solve the approximated problem in equation \eqref{eq:approx-wssd} under the epipolar constraint introduced in equation \eqref{eq:epipolar-constraint}:
\begin{align}
	\widetilde{Q}_{k}(\vec{v}_{k}) &= \vec{v}_{k}^{T} \cdot \mat{A}_{k} \cdot \vec{v}_{k} + 2 \cdot \vec{v}_{k}^{T} \cdot \vec{b}_{k} + c_{k}
	\nonumber
	\\
	&\phantom{=} + 2 \cdot \lambda \cdot \left( (\vec{y}_{k}^{T} + \vec{v}^{T}, 1) \cdot \mat{F} \cdot 
	\begin{pmatrix}
		\vec{x}_{k}	\\
		1
	\end{pmatrix}
	\right).
	\label{eq:constrained-approx-wssd}
\end{align}
The optimal displacement $\vec{v}_{k, \textnormal{opt}}$ can be computed via the minimization of $\widetilde{Q}_{k}(\vec{v}_{k})$:
\begin{align}
	&& \frac{1}{2} \cdot \frac{\partial \widetilde{Q}_{k}}{\partial \vec{v}_{k}}(\vec{v}_{k, \textnormal{opt}}) = \mat{A}_{k} \cdot \vec{v}_{k, \textnormal{opt}} + \vec{b}_{k} + \lambda \cdot \mat{F}^{'} \cdot
	\begin{pmatrix}
		\vec{x}_{k}	\\
		1
	\end{pmatrix}
	&\stackrel{!}{=} \vec{0}.
\end{align}
In combination with equation \eqref{eq:epipolar-constraint} this yields the linear equation system:
\begin{align}
	\nonumber
	\begin{pmatrix}
		\mat{A}_{k}
		&
		\mat{F}^{'} \cdot
		\begin{pmatrix}
			\vec{x}_{k}	\\
			1
		\end{pmatrix}
		\\
		\left( \mat{F}^{'} \cdot
		\begin{pmatrix}
			\vec{x}_{k}	\\
			1
		\end{pmatrix}
		\right)^{T}
		&
		0									
	\end{pmatrix}
	\cdot
	\begin{pmatrix}
		\vec{v}_{k, \textnormal{opt}}	\\
		\lambda
	\end{pmatrix}
	\stackrel{!}{=}
	\\
	\begin{pmatrix}
		-\vec{b}_{k}	\\
		-(\vec{y}_{k}^{T}, 1) \cdot \mat{F} \cdot
		\begin{pmatrix}
			\vec{x}_{k}	\\
			1
		\end{pmatrix}
	\end{pmatrix}.
	\label{eq:lk-displacement-constrained-les}
\end{align}
The matrix on the left hand side of this equation system is symmetric and $\mat{F}$ is a function of the motion parameter $\vec{p}$, i.e. there is a closed form solution to the optimal displacement $\vec{v}_{k, \textnormal{opt}}$ showing the following dependency:
\begin{align}
	\vec{v}_{k, \textnormal{opt}} = \vec{f}(\mat{A}_{k}, \vec{b}_{k}, \vec{x}_{k}, \vec{y}_{k}, \mat{K}, \vec{p}) = \vec{f}_{k}(\vec{p}).
	\label{eq:displacement-dependencies}
\end{align}
For a given epipolar geometry (which is equivalent to a given unscaled relative pose and a calibrated camera) the linear equation system \eqref{eq:lk-displacement-constrained-les} is the extension of the standard Lucas-Kanade equation (see equation \eqref{eq:approx-wssd}) with an epipolar constraint. It can be used to optimize image correspondences if the epipolar geometry is already known in beforehand.

\subsection{Joint Epipolar Tracking}
\label{sec:joint-epipolar-tracking}

The present work extends the epipolar constrained tracking in the following sense: We do not only optimize each feature correspondence $\vec{x}_{k} \leftrightarrow \vec{y}_{k}$ individually with respect to a given epipolar geometry, but build a joint loss function which can be optimized with respect to the underlying motion that characterizes the displacements of \emph{all} image points (given that all points obey the same epipolar relation).

Using this approach, we can additionally optimize the motion parameters themselves. We call this method \emph{Joint Epipolar Tracking} (JET). To this end, we perform a re-parametrization of the loss function $\widetilde{Q}_{k}(\vec{v}_{k})$ by substituting the functional dependency $\vec{v}_{k}=\vec{f}_{k}(\vec{p})$ into it (compare equations \eqref{eq:constrained-approx-wssd} and \eqref{eq:displacement-dependencies} respectively):
\begin{align}
	\widetilde{Q}_{k}(\vec{p}) &\stackrel{def}{=} \widetilde{Q}_{k}(\vec{v} = \vec{f}_{k}(\vec{p}))
	\nonumber
	\\
	&\stackrel{\phantom{def}}{=} \vec{f}_{k}^{T}(\vec{p}) \cdot \mat{A}_{k} \cdot \vec{f}_{k}(\vec{p}) + 2 \cdot \vec{f}_{k}^{T}(\vec{p}) \cdot \vec{b}_{k} + c_{k}
	\nonumber
	\\
	&\phantom{=} + 2 \cdot \lambda \cdot \underbrace{\left( (\vec{y}_{k}^{T} + \vec{f}_{k}^{T}(\vec{p}), 1) \cdot \mat{F} \cdot
	\begin{pmatrix}
		\vec{x}_{k}	\\
		1
	\end{pmatrix}
	\right)}_{0}.
	\label{eq:single-ssd}
\end{align}
By using this definition of the displacement, the optimization of the loss function is no longer performed with respect to an image displacement $\vec{v}_{k}$ but with respect to an epipolar geometry which is induced by the relative pose of the camera and the environment. This relative pose is evoking the optical flow in the image domain.

Joining together the loss functions from equation \eqref{eq:single-ssd} for several feature correspondences $\left\{ \vec{x}_{k} \leftrightarrow \vec{y}_{k} \right\}_{k}$ and adding a prior term for the motion (expressing a statistical model of `typical` motion) yields the joint loss function
\begin{align}
	\widetilde{Q}(\vec{p}) &= \underbrace{\frac{1}{N} \sum_{k = 1}^{N} \widetilde{Q}_{k}(\vec{p})}_{\textnormal{image information}} \quad + \quad \xi_{Q} \cdot \underbrace{(\vec{p} - \hat{\vec{p}})^{T} \cdot \mat{C}_{\vec{p} - \hat{\vec{p}}}^{-1} \cdot (\vec{p} - \hat{\vec{p}})}_{\textnormal{prior term on motion parameters}}.
	\label{eq:joint-wssd-with-prior}
\end{align}
The minimization of this function allows to determine the motion parameters, and hence the unscaled relative pose, that best describes the optical flow. In equation \eqref{eq:joint-wssd-with-prior} the part of the joint loss function that is dependent on the image information has been extended by a second part that incorporates statistical prior knowledge coupled via the coupling constant $\xi_{Q}$. The prior information is characterized by a prediction of the expected motion $\hat{\vec{p}}$ and a covariance matrix of the prediction residuals $\mat{C}_{\vec{p} - \hat{\vec{p}}}$.

These motion prior terms are determined by a linear regression approach on a dataset of motion parameters that are representative for the type of motion to be expected (e.g. restricted car motion, unrestricted motion of a handheld camera). We use a very similar approach as in \cite{BradlerWiegandMester2015ICCV-CVRSUAD} to determine the parameters of a linear predictor. The difference is that we employ a third order predictor, i.e. the preceding three motion parameter sets are taken into consideration when evaluating the statistics and performing the dynamic prediction. 

Equation \eqref{eq:joint-wssd-with-prior} can be expressed in vertex form and the optimization of $\widetilde{Q}(\vec{p})$ is represented as the following least squares problem:
\begin{align}
	\nonumber
	\widetilde{Q}(\vec{p}) = \sum_{k = 1}^{N} \| \vec{q}_{k}(\vec{p}) \|_{2}^{2} + \textnormal{const.} \quad ,
	\\
	\vec{q}_{k}(\vec{p}) = \sqrt{\frac{1}{N}} \cdot
	\begin{pmatrix}
		\mat{A}_{k}^{1/2} \cdot (\vec{f}_{k}(\vec{p}) + \mat{A}_{k}^{-1} \cdot \vec{b}_{k})
		\\
		\sqrt{\xi_{Q}} \cdot \mat{C}_{\vec{p} - \hat{\vec{p}}}^{-1/2} \cdot (\vec{p} - \hat{\vec{p}})
	\end{pmatrix}.
\end{align}
With an initial estimate $\vec{p}^{(0)}$ of the motion parameters (\eg the prediction $\hat{\vec{p}}$ based on the previous motion parameters), we can now solve this optimization problem using a nonlinear solver like the Ceres solver \cite{ceres-solver}. The result of this motion optimization $\vec{p}_{\textnormal{opt}}$ is then used to improve the feature correspondences $\vec{x}_{k} \leftrightarrow \vec{y}_{k}$ by shifting the corresponding image point to its epipolar line by $\vec{y}_{k} \rightarrow \vec{y}_{k} + \vec{f}_{k}(\vec{p}_{\textnormal{opt}})$ (see equations \eqref{eq:lk-displacement-constrained-les} and \eqref{eq:displacement-dependencies}). Since the original image difference has been replaced by a linear approximation during the Gauss-Newton approach at the beginning, these improved correspondences and the improved motion parameters serve as an initialization for the second iteration step of this optimization procedure. We continue with this procedure as long as the target loss function
\begin{align}
	Q(\vec{p}_{\textnormal{opt}}) \definedas \frac{1}{N} \sum_{k = 1}^{N} Q_{k}(\vec{v}_{k} = \vec{f}_{k}(\vec{p}_{\textnormal{opt}}))
	\label{eq:joint-wssd}
\end{align}
is decreased. Note that the target loss function incorporates the \emph{exact} image difference as introduced in equation \eqref{eq:exact-wssd}.

The optimization of the relative pose using JET does not merely minimize the reprojection error\footnote{Actually the reprojection error is zero, since the feature correspondences are just optimized with respect to the relative pose.}, but rather than that minimize the photometric error of the feature correspondences by including the full image information encoded in the quantities $\mat{A}_{k}$, $\vec{b}_{k}$ and $c_{k}$. Compared to other leading direct methods, such as \cite{EngelKoltunCremers2016,EngelSchoepsCremers2014}, JET is the most compact formulation of the direct 2-view $n$ points pose optimization problem based on minimizing the photometric error.
\section{Experiments}

\newcolumntype{C}[1]{>{\centering\arraybackslash}p{#1}}
\begin{table*}[htb!]
	\centering
	\begin{tabular}{|l||C{0.7cm}|C{0.7cm}|C{0.7cm}|C{0.7cm}|C{0.7cm}|C{0.7cm}||C{0.7cm}|C{0.7cm}|C{0.7cm}|C{0.7cm}|C{0.7cm}|C{0.7cm}|}
		\hline
		\multirow{2}{*}{\textbf{Dataset}} & \multicolumn{2}{c|}{$\boldsymbol{\rho}_{in}$}  & \multicolumn{2}{c|}{$\boldsymbol{\rho}_{JET}$} & \multicolumn{2}{c||}{$\boldsymbol{\rho}_{RPE}$}  & \multicolumn{2}{c|}{$\boldsymbol{\Omega}_{in}$} & \multicolumn{2}{c|}{$\boldsymbol{\Omega}_{JET}$} & \multicolumn{2}{c|}{$\boldsymbol{\Omega}_{RPE}$} \\ \cline{2-13}
		& $\mu$ & $\sigma$ & $\mu$ & $\sigma$ & $\mu$ & $\sigma$ & $\mu$ & $\sigma$ & $\mu$ & $\sigma$ & $\mu$ & $\sigma$ \\ 
		\hline
		\hline
		Construction & 0.96 & 0.28 & \bf{0.06} & 0.09 & 0.11 & 0.08  & 7.65 & 2.83 & \bf{1.62} & 2.39 & 2.04 & 1.71 \\
		Construction* & 0.96 & 0.28 & \bf{0.03} & 0.03 & 0.1 & 0.06 & 7.64 & 2.84 & \bf{0.77} & 0.87 & 0.89 & 0.99 \\
		Highway & 0.96 & 0.28 & \bf{0.03} & 0.05 & 0.06 & 0.03 & 7.63 & 2.85 & \bf{0.6} & 1.4 & 1.12 & 0.91 \\
		Highway* & 0.96 & 0.28 & \bf{0.02} & 0.02 & 0.05 & 0.05 & 7.65 & 2.85 & \bf{0.25} & 0.47 & 0.23 & 0.18 \\
		LivingRoom02 & 4.8 & 1.39 & \bf{0.38} & 0.39 & 0.64 & 0.47 & \bf{12.9} & 5.61 & 14.8 & 14.7 & 20.1 & 14.8 \\
		OfficeRoom02 & 4.81 & 1.39 & \bf{0.41} & 0.49 & 0.71 & 0.72 & \bf{12.8} & 5.62 & 18.9 & 20.7 & 23.1 & 18.0 \\
		\hline
	\end{tabular}
	\caption{First two moments of $\rho$ and $\Omega$ from the evaluation on COnGRATS and ICL-NUIM RGB-D dataset. Datasets ending with * indicate the use of prior knowledge. All values are in degrees.}
	\label{tab:mean-and-standard-deviation-rho-omega}
\end{table*}
We evaluated the JET procedure presented here on synthetic data \cite{BiedermannOchsMester2015IVCNZ,HandaWhelanMcDonaldDavison2014ICRA}, applying noise to the different input parameters to investigate the stability against noise in our components. As we used synthetic data, we had perfect ground truth for our results to compare against, a situation usually very hard to obtain for real-life driving scenarios, \eg\cite{GeigerLenzUrtasun2012CVPR,GeigerLenzStillerUrtasun2013IJRR}.

The aim in our experiment is to optimize the motion parameters and correct the feature correspondences $\{\vec{x}_{k} \leftrightarrow \vec{y}_{k}\} \rightarrow \{\vec{x}_{k} \leftrightarrow \vec{y}_{k, \textnormal{opt}}\}$ so that they obey the epipolar geometry induced by the optimized motion parameters $\vec{p}_{\textnormal{opt}}$:
\begin{align}
	\begin{pmatrix}
		\vec{y}_{k, \textnormal{opt}}	\\
		1
	\end{pmatrix}
	^{T} \cdot \mat{F}(\vec{p}_{\textnormal{opt}}) \cdot
	\begin{pmatrix}
		\vec{x}_{k} \\
		1
	\end{pmatrix}
	= 0 \quad \textnormal{, $k \in \{1,\ldots, N\}$}
\end{align}
We compare the results achieved with JET against the results achieved with a method that minimizes the reprojection error (RPE).

\subsection{Competing method: Optimization of the reprojection error}

The competitor \emph{RPE optimization} is a method that minimizes the reprojection error and performs the following steps:
\begin{enumerate}
	\item Optimize correspondences: \\
	$\{\vec{x}_{k} \leftrightarrow \vec{y}_{k}\}
	 \rightarrow \{\vec{x}_{k} \leftrightarrow \vec{y}_{k}^{'}\}$ (optional)
	\item Minimize the reprojection error: \\
	$\vec{p} \rightarrow \vec{p}_{\textnormal{opt}}$
	\item Perform a minimum correction of the correspondences,
	 so that they are in agreement with $\vec{p}_{\textnormal{opt}}$: \\
	 $\{\vec{x}_{k} \leftrightarrow \vec{y}_{k}^{'}\}
	 \rightarrow \{\vec{x}_{k} \leftrightarrow \vec{y}_{k, \textnormal{opt}}\}$
\end{enumerate}
The first task is optional and optimizes the feature correspondences using standard Lucas-Kanade tracking as it is implemented in OpenCV \cite{Opencv}. We will run experiments with both, step one enabled and disabled. In the mandatory second step, RPE optimizes the motion parameters by minimizing the reprojection error
\begin{align}
	d_{k}(\vec{p}) \definedas
	\begin{pmatrix}
		\vec{y}_{k}	\\
		1
	\end{pmatrix}
	^{T} \cdot \mat{F}(\vec{p}) \cdot
	\begin{pmatrix}
		\vec{x}_{k} \\
		1
	\end{pmatrix}
	/ \left\| \mat{F}^{'}(\vec{p}) 
	\begin{pmatrix}
		\vec{x}_{k} \\
		1
	\end{pmatrix}
	\right\|_{2}
\end{align}
for all feature correspondences. $d_{k}(\vec{p})$ is the distance of the image point $\vec{y}_{k}$ to the epipolar line specified by the fundamental matrix $\mat{F}(\vec{p})$ (see equation \eqref{eq:fundamental-matrix-definition}) and $\vec{x}_{k}$. We delegate the optimization of the loss function $R(\vec{p})$ of the RPE method
to the \emph{Ceres-Solver} from Google \cite{ceres-solver}.
\begin{align}
	\nonumber
	R(\vec{p}) &\definedas \frac{1}{N} \cdot \sum_{k = 1}^{N} d_{k}^{2}(\vec{p}) \quad
	+ \quad \xi_{R} \cdot (\vec{p} - \hat{\vec{p}})^{T}
	\cdot \mat{C}_{\vec{p} - \hat{\vec{p}}}^{-1} \cdot (\vec{p} - \hat{\vec{p}})
	\\
	\nonumber
	&= \sum_{k = 1}^{N} \left\| \vec{r}_{k}(\vec{p}) \right\|_{2}^{2} \quad , \textnormal{with} \\
	\vec{r}_{k}(\vec{p}) &= \sqrt{\frac{1}{N}} \cdot
	\begin{pmatrix}
		d_{k}(\vec{p})	\\
		\sqrt{\xi_{R}} \cdot \mat{C}_{\vec{p} - \hat{\vec{p}}}^{-1/2} \cdot (\vec{p} - \hat{\vec{p}})
	\end{pmatrix}
\end{align}
After having computed the optimized motion parameters $\vec{p}_{\textnormal{opt}}$, we determine the optimized corresponding points $\vec{y}_{k, \textnormal{opt}}$ by projecting all $\vec{y}_{k}$ to the closest points on their respective epipolar line. For that purpose we introduce the abbreviations
\begin{align}
	\vec{l}_{k} =
	\begin{pmatrix}
		l_{k, 0}	\\
		l_{k, 1}	\\
		l_{k, 2}
	\end{pmatrix}
	= \mat{F}(\vec{p}_{\textnormal{opt}}) \cdot
	\begin{pmatrix}
		\vec{x}_{k}	\\
		1
	\end{pmatrix}
	\quad , \quad \vec{y}_{k} =
	\begin{pmatrix}
		y_{k, 0}	\\
		y_{k, 1}
	\end{pmatrix}
\end{align}
and obtain for the optimized corresponding point:
\begin{align}
	\nonumber
	\vec{y}_{k, \textnormal{opt}} &= \frac{1}{(l_{k, 0})^{2} + (l_{k, 1})^{2}} \cdot
	\\
	&
	\begin{pmatrix}
		y_{k, 0} \cdot (l_{k, 1})^{2} - y_{k, 1} \cdot l_{k, 0} \cdot l_{k, 1} -l_{k, 0} \cdot l_{k, 1}
		\\
		-y_{k, 0} \cdot l_{k, 0} \cdot l_{k, 1} + y_{k, 1} \cdot (l_{k, 0})^{2} -l_{k, 1} \cdot l_{k, 2}
	\end{pmatrix}.
\end{align}

\subsection{Initialization}

Both methods were initialized with exactly the same estimated image correspondences and the same estimate of motion parameters. When using synthetic data, it is straightforward to obtain ground truth reference values for the correspondences as well as for the motion parameters. The COnGRATS \cite{BiedermannOchsMester2015IVCNZ} scenes we used in the evaluation, re-use pose sequences from the KITTI Benchmark. To make a coarse estimate of 
the variation range of the motion parameters, we checked the statistics of the motion parameters on the KITTI dataset, which covers a wide range of driving scenarios and can be considered as representative for realistic car motion.

If we assume a normal distribution of $\vec{p}_{n} - \vec{p}_{n - 1}$ and use the KITTI motion statistic to find upper bounds for the variances of the translational and rotational degrees of freedom ($\sigma_{\textnormal{rot}}^{2} < 10^{-5}$ and $\sigma_{\textnormal{trans}}^{2} < 10^{-3}$), we can estimate the $3 \, \sigma$ interval to be $3 \, \sigma_{\textnormal{rot}} < 10^{-2}$ and $3 \, \sigma_{\textnormal{trans}}< 10^{-1}$. More than $99.7$\% of the motion parameters do not deviate by more than $3 \, \sigma_{\textnormal{rot/trans}}$ from their temporal predecessor.

We use these insights to justify a realistic variation range of $\pm1^{\circ}$ and $\pm10^{\circ}$ for the rotation and translation parameters respectively. These ranges correspond to more than $5$ standard deviations $\sigma_{\textnormal{rot/trans}}$.
We apply uniformly distributed noise with the just derived intervals to the motion parameters.

A similar consideration for a hand held camera, as it is used in the second synthetic dataset \cite{HandaWhelanMcDonaldDavison2014ICRA}, leads to a variation range of $\pm5^{\circ}$ and $\pm20^{\circ}$ for the rotation and translation parameters, respectively.

For the corresponding image points $\vec{y}_{k}$, we apply uniform noise to the $x$- and $y$-component of the ground truth value, each with a level of $\pm5$ pixels.

\subsection{Evaluation measures}

\begin{table*}[htb!]
\begin{center}
\begin{tabular}{|c||C{1.5cm}|C{1.5cm}||C{1.5cm}|C{1.5cm}||C{1.5cm}|C{1.5cm}|}
\hline
\multirow{2}{*}{\textbf{\begin{tabular}[c]{@{}c@{}}KITTI\\ Seq No.\end{tabular}}} & \multicolumn{2}{c|}{\textbf{Rotation $\boldsymbol{\rho}$ [\textit{deg}]}} & \multicolumn{2}{c|}{\textbf{Translation $\boldsymbol{\Omega}$ [\textit{deg}]}} & \multicolumn{2}{c|}{\textbf{SSD}} \\ \cline{2-7}                                                                                   & \textbf{RPE}       & \textbf{JET}       & \textbf{RPE}         & \textbf{JET}         & \textbf{RPE}    & \textbf{JET}    \\
\hline
\hline
0  &  0.188  &  \textbf{0.096}  &  \textbf{6.582}  &  6.749  &  1209.95  &  \textbf{180.39}  \\ \hline
1  &  0.364  &  \textbf{0.253}  &  8.374  &  \textbf{7.998}  &  1553.32  &  \textbf{178.64}  \\ \hline
2  &  0.154  &  \textbf{0.061}  &  1.703  &  \textbf{1.502}  &  1178.16  &  \textbf{266.20}  \\ \hline
3  &  0.098  &  \textbf{0.035}  &  0.970  &  \textbf{0.891}  &  388.74  &  \textbf{138.38}  \\ \hline
4  &  0.142  &  \textbf{0.045}  &  1.123  &  \textbf{0.951}  &  774.49  &  \textbf{176.47}  \\ \hline
5  &  0.138  &  \textbf{0.049}  &  1.445  &  \textbf{1.274}  &  1120.63  &  \textbf{212.27}  \\ \hline
6  &  0.436  &  \textbf{0.358}  &  11.129  &  \textbf{10.765}  &  1342.75  &  \textbf{220.85}  \\ \hline
7  &  0.262  &  \textbf{0.152}  &  28.411  &  \textbf{28.070}  &  1587.19  &  \textbf{219.49}  \\ \hline
8  &  0.169  &  \textbf{0.063}  &  9.536  &  \textbf{9.429}  &  1188.83  &  \textbf{206.49}  \\ \hline
9  &  0.107  &  \textbf{0.029}  &  0.752  &  \textbf{0.676}  &  687.81  &  \textbf{243.71}  \\ \hline
10  &  0.237  &  \textbf{0.131}  &  1.614  &  \textbf{1.361}  &  1218.97  &  \textbf{275.59}  \\ \hline
\end{tabular}
\end{center}
\caption{Evaluation on the KITTI training dataset.}
\label{tab:result-kitti}
\end{table*}
Each experiment gets initialized with an approximation of the pose and with initial image correspondences. To quantify the quality of the input and the output of the methods, the deviation from ground truth is expressed by the following four evaluation measures:
\begin{itemize}
	\item \emph{Rodrigues angle $\rho$ (rotational error):} \\
The rotation parameters $\theta$, $\psi$ and $\phi$ define a rotation matrix $\mat{R}$ which is to be compared against the ground truth $\mat{R}_{\textnormal{gt}}$ via the relative rotation $\mat{R}_{\textnormal{rel}} = \mat{R}_{\textnormal{gt}} \cdot \mat{R}^{T}$. According to Rodrigues` formula, $\mat{R}_{\textnormal{rel}}$ can be interpreted as a rotation of an angle $\rho$ about some axis $\vec{n}$.
The absolute value of the Rodrigues angle $\rho$ serves as a measure for the deviation from the ground truth rotation.

	\item \emph{Angle of intersection $\Omega$ (translational error):} \\
The translation parameters $\alpha$ and $\beta$ represent the direction of the translation vector. The translation direction is compared to the ground truth via the absolute value of its angle of intersection $\Omega$.

	\item \emph{RMS distance of corresponding points (positional error):} \\
The quality of the point correspondences is characterized by the mean deviation from ground truth: $\textnormal{RMS} \definedas \sqrt{\frac{1}{N} \sum_{k = 1}^{N} \| \vec{y}_{k} - \vec{y}_{k, \textnormal{gt}} \|_{2}^{2}}$.
	
	\item \emph{Joint weighted sum of squared differences SSD (photometric error):} \\
	The only measure that is absolute and not relative to the ground truth is the SSD. It is the average squared gray value difference over all patches of the image correspondences $\vec{x}_{k} \leftrightarrow \vec{y}_{k} = \vec{x}_{k} + \vec{v}_{k}$: \\
$Q \definedas \frac{1}{N} \sum_{k = 1}^{N} \sum_{\vec{x}} \mat{W}[\vec{x}- \vec{x}_{k}] \cdot \left( \imgI[\vec{x}] - \imgJ[\vec{x} + \vec{v}_{k}] \right)^{2}$.
\end{itemize}

\subsection{COnGRATS \& ICL-NUIM RGB-D dataset}

\begin{figure*}[ht]
	\centering
	\includegraphics[width=.33\linewidth]{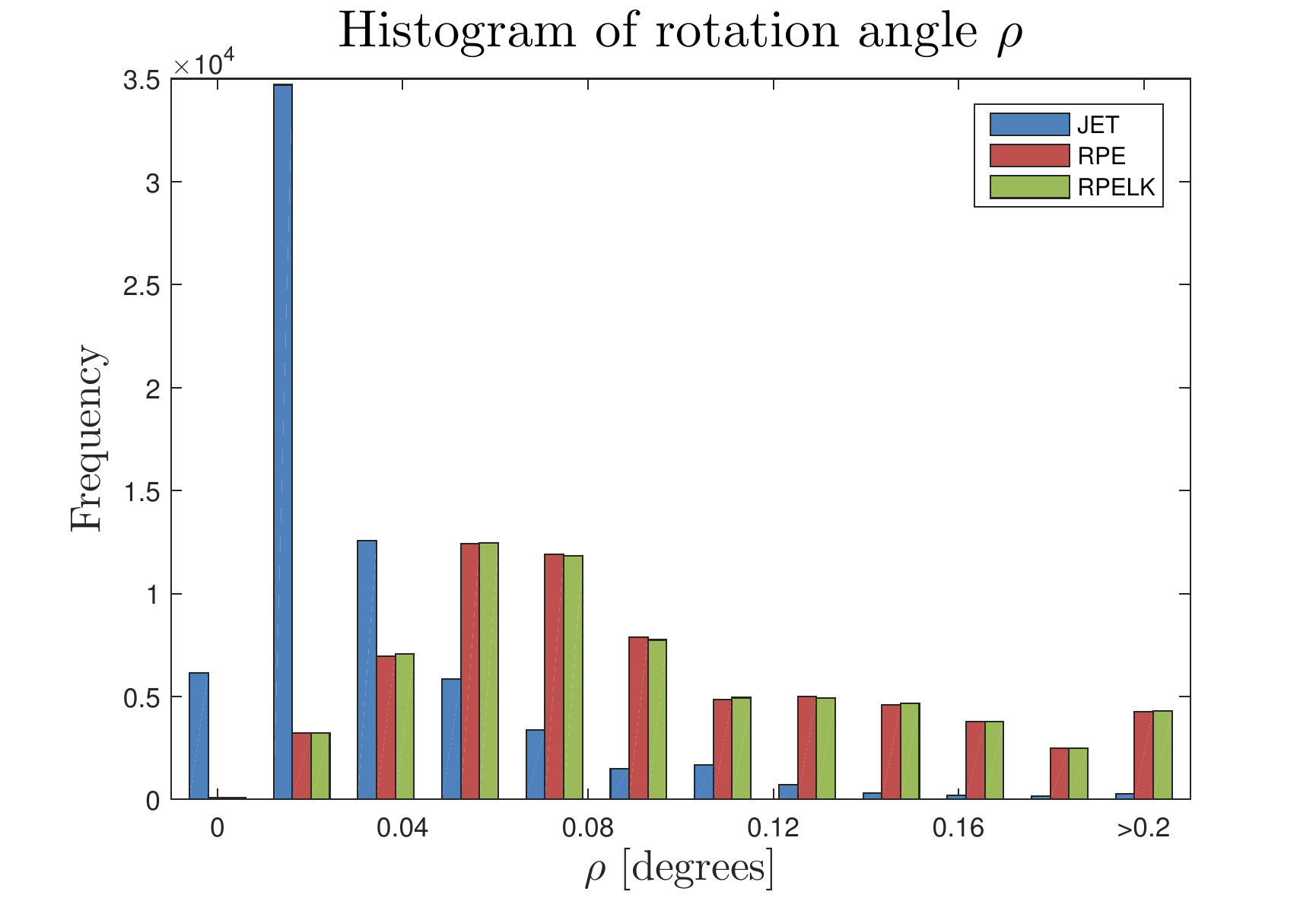}
	\includegraphics[width=.33\linewidth]{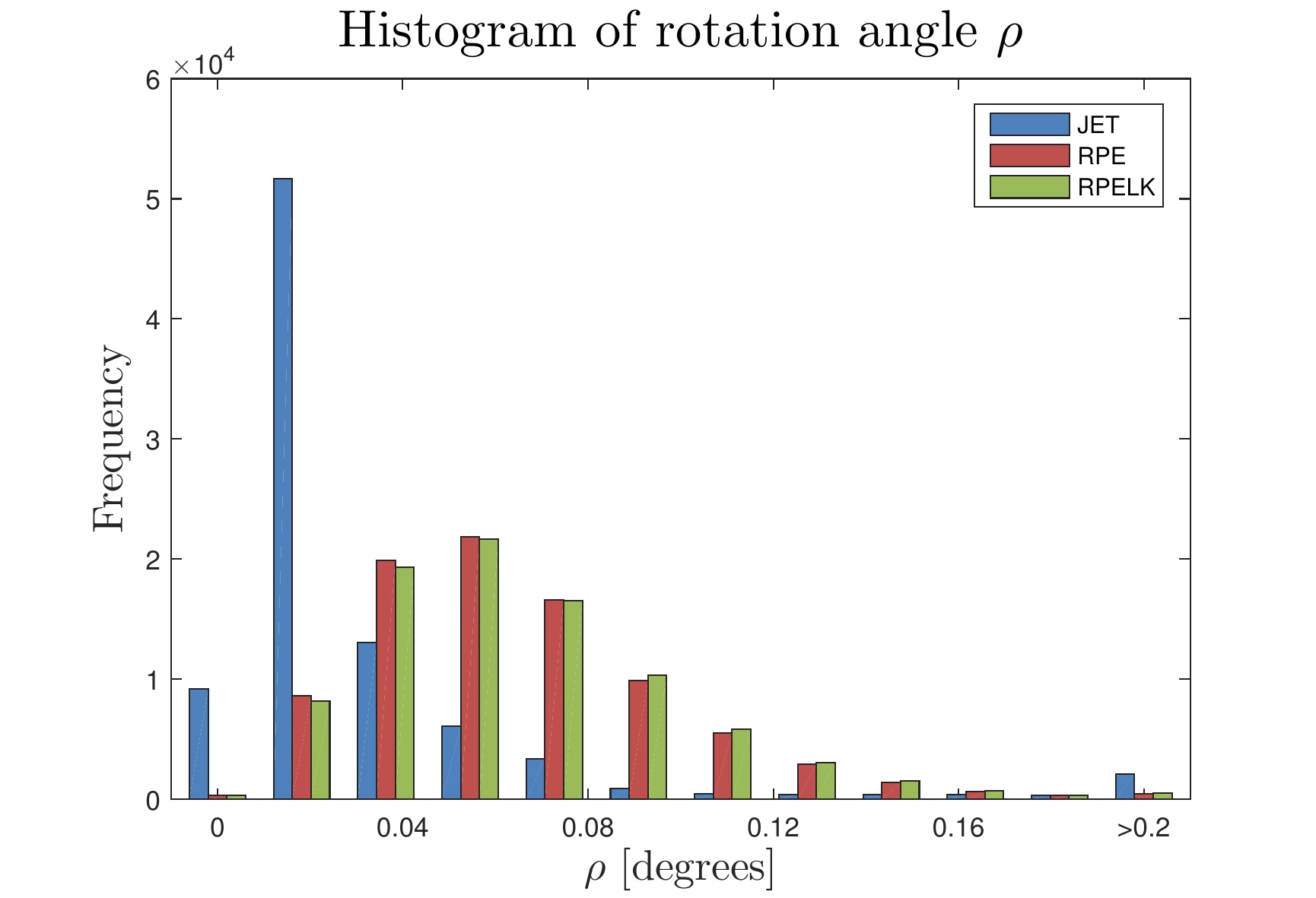}
	\includegraphics[width=.33\linewidth]{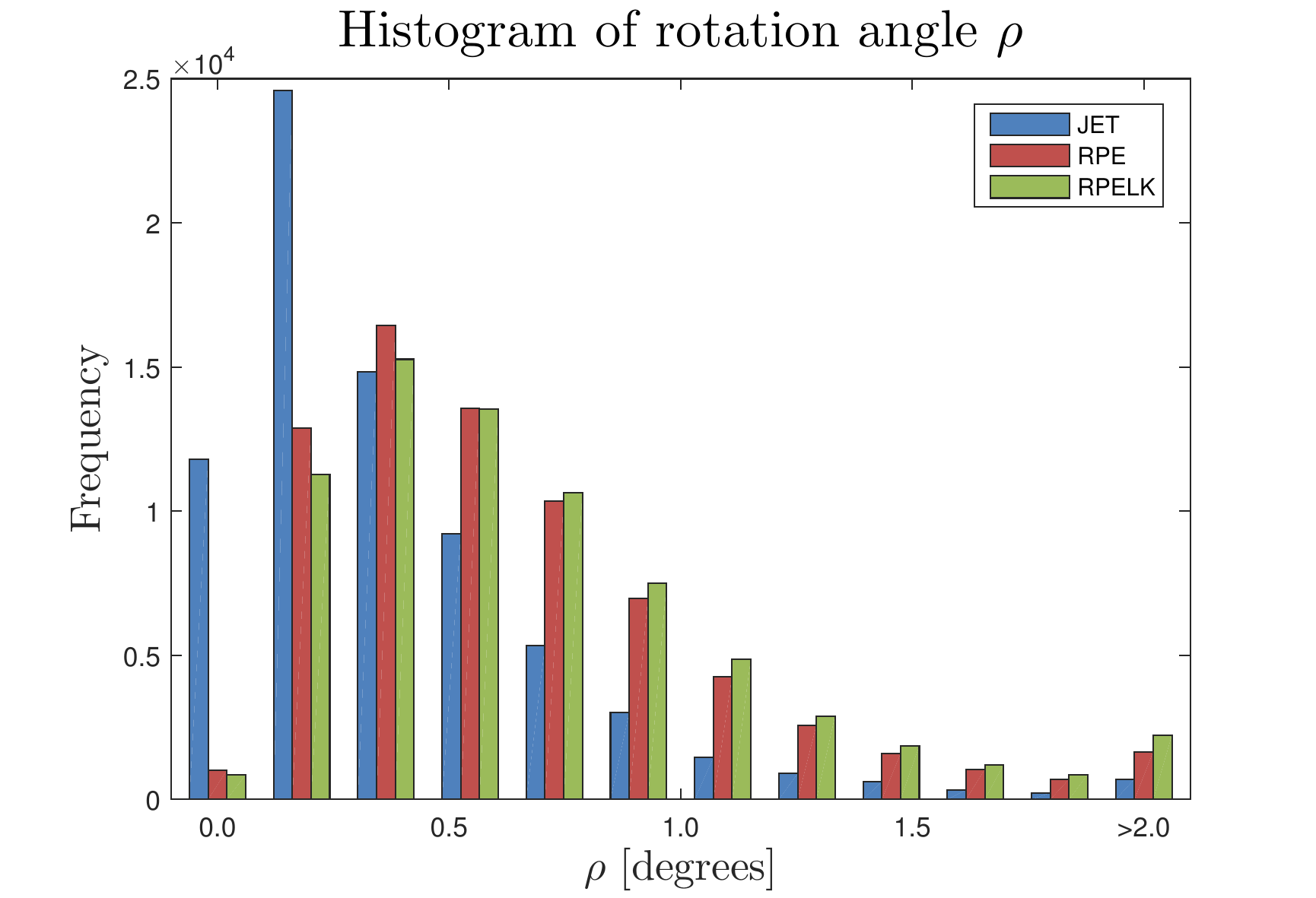}
	\includegraphics[width=.33\linewidth]{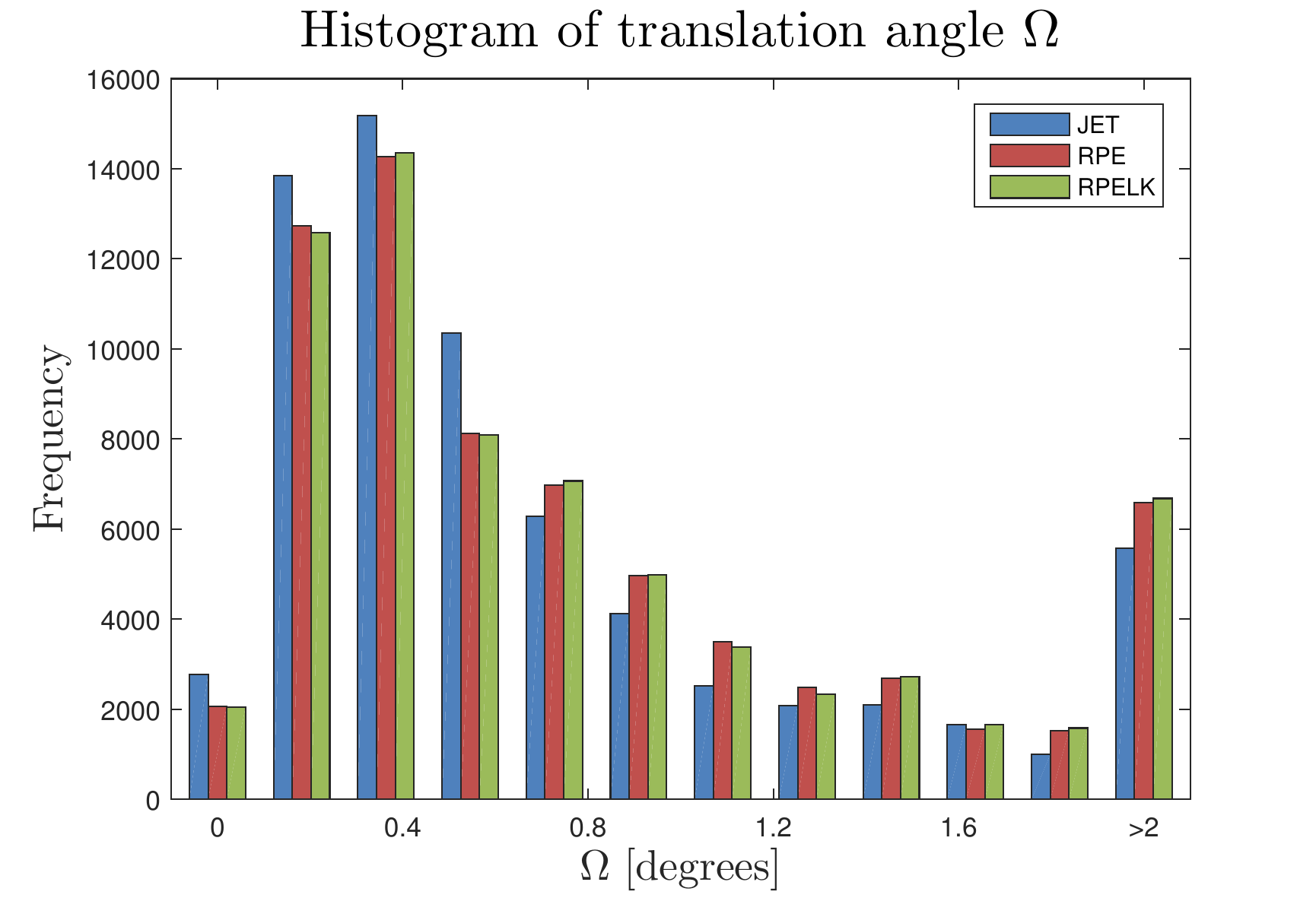}
	\includegraphics[width=.33\linewidth]{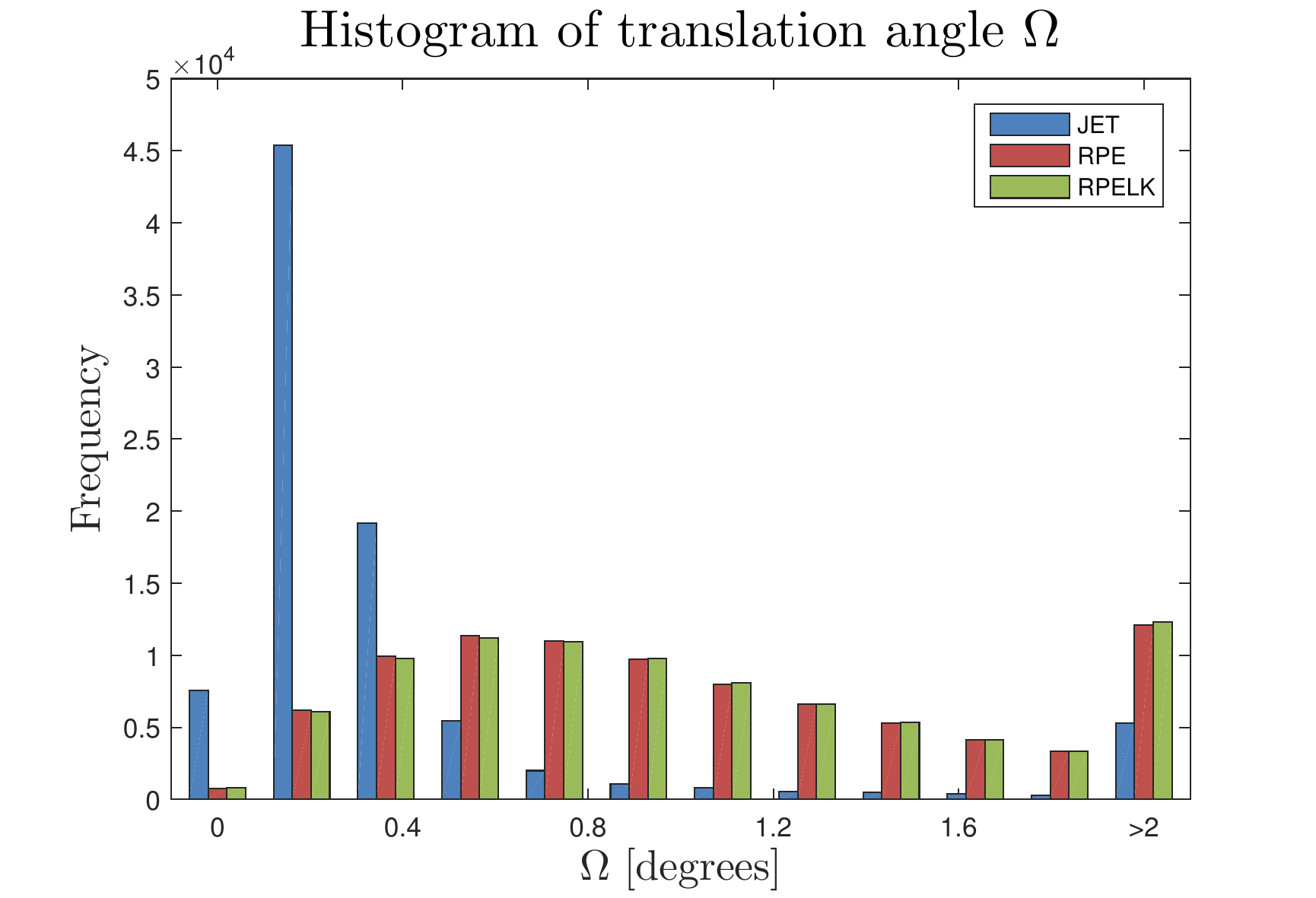}
	\includegraphics[width=.33\linewidth]{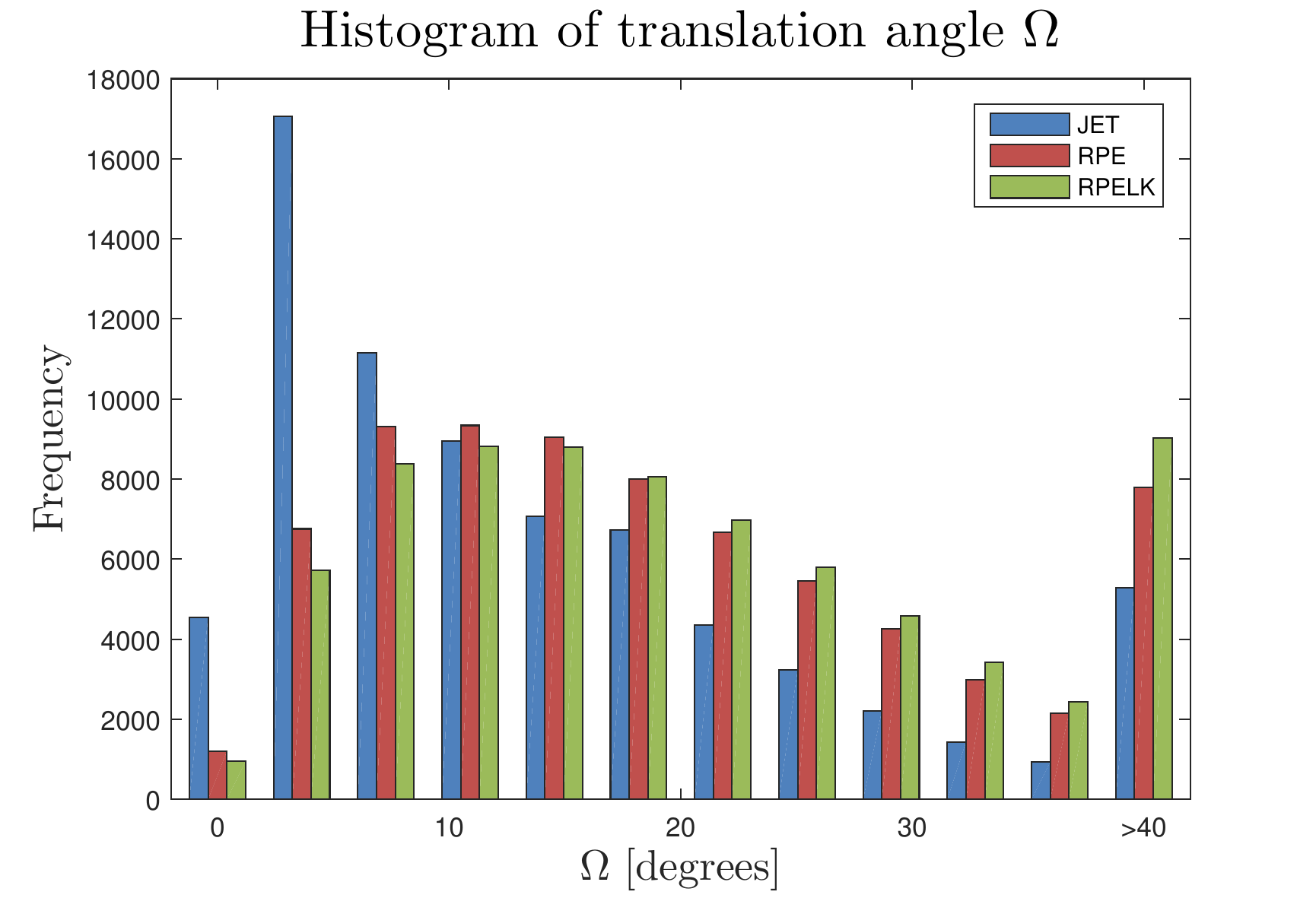}
	\includegraphics[width=.33\linewidth]{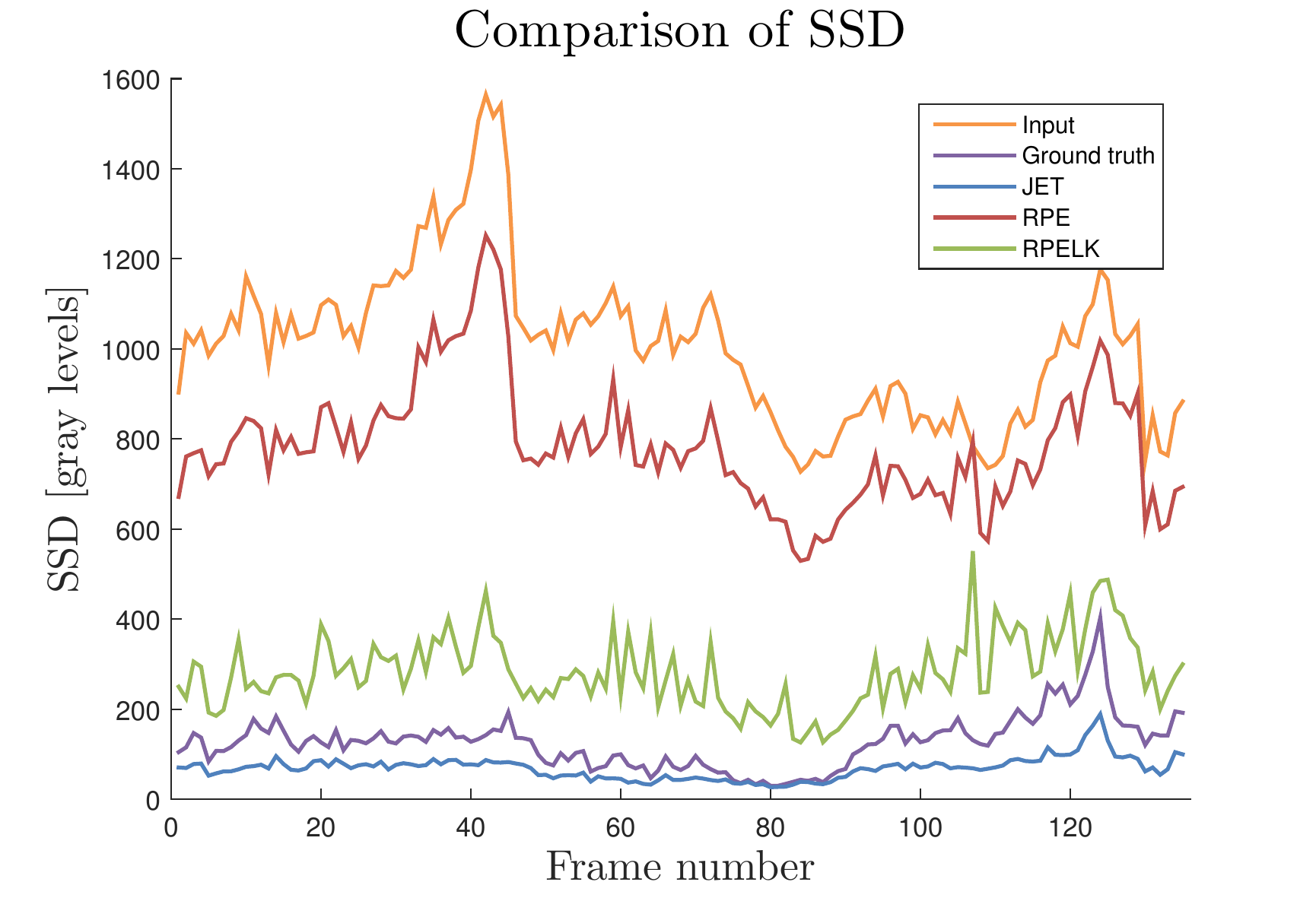}
	\includegraphics[width=.33\linewidth]{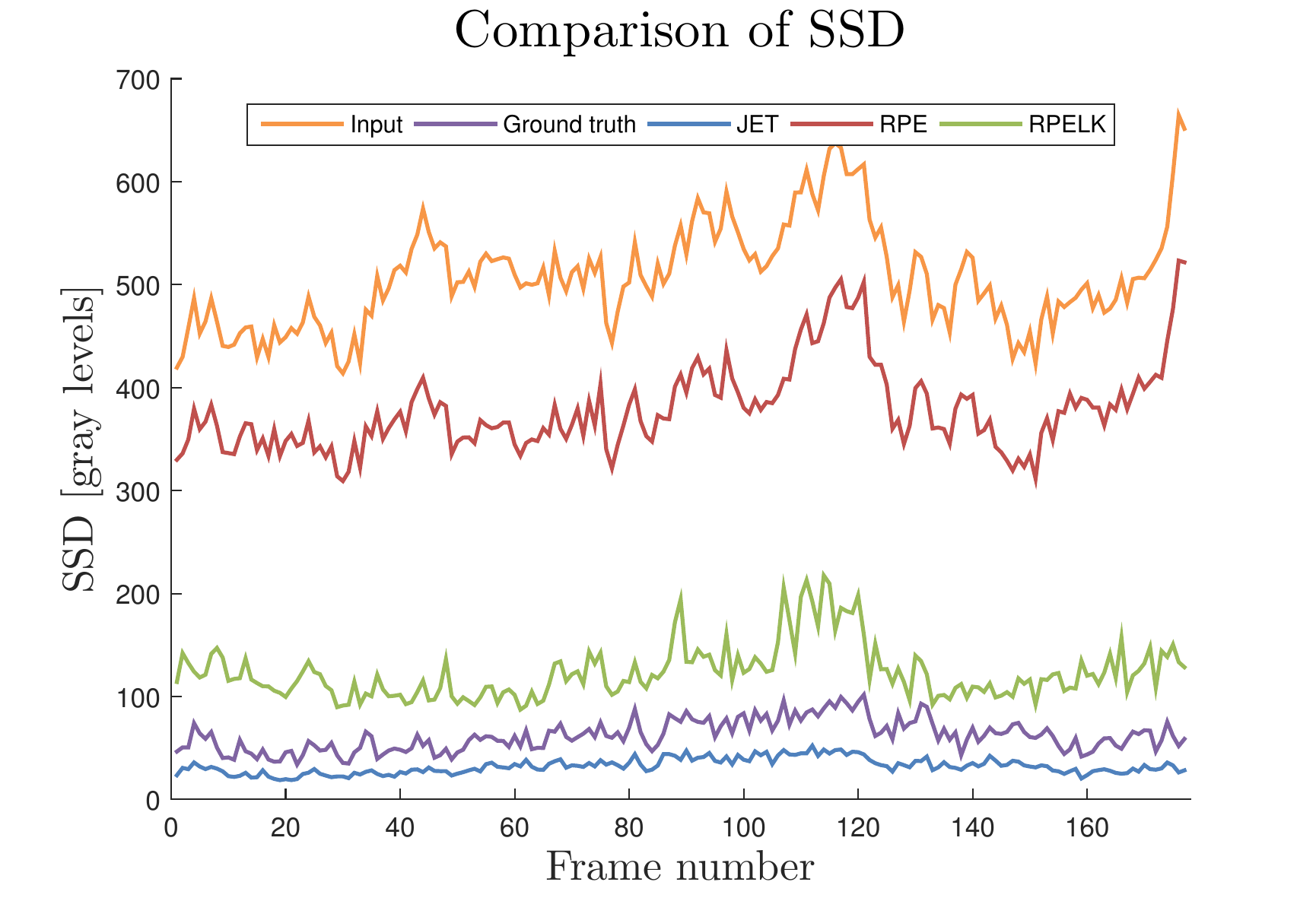}
	\includegraphics[width=.33\linewidth]{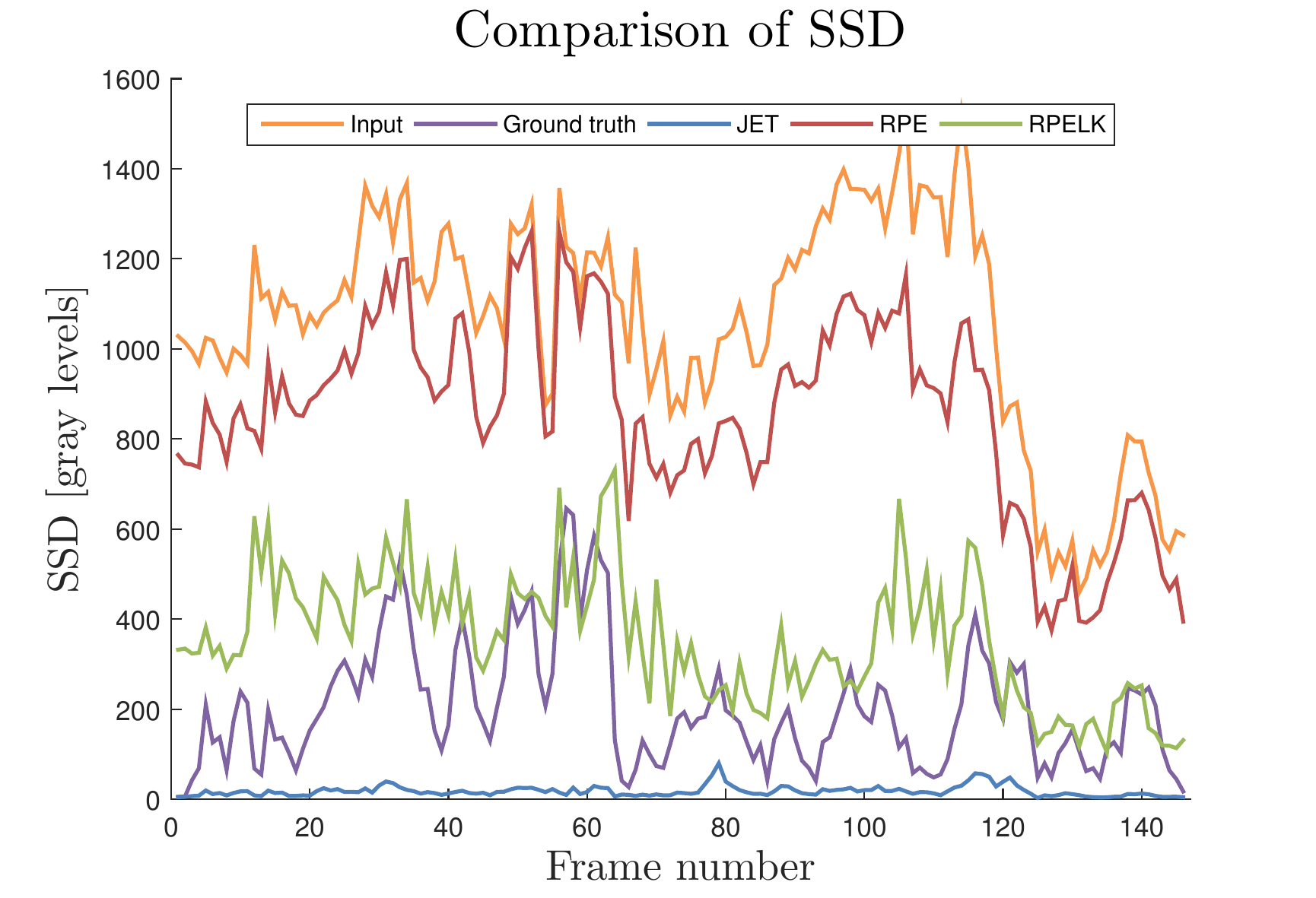}
	\includegraphics[width=.33\linewidth]{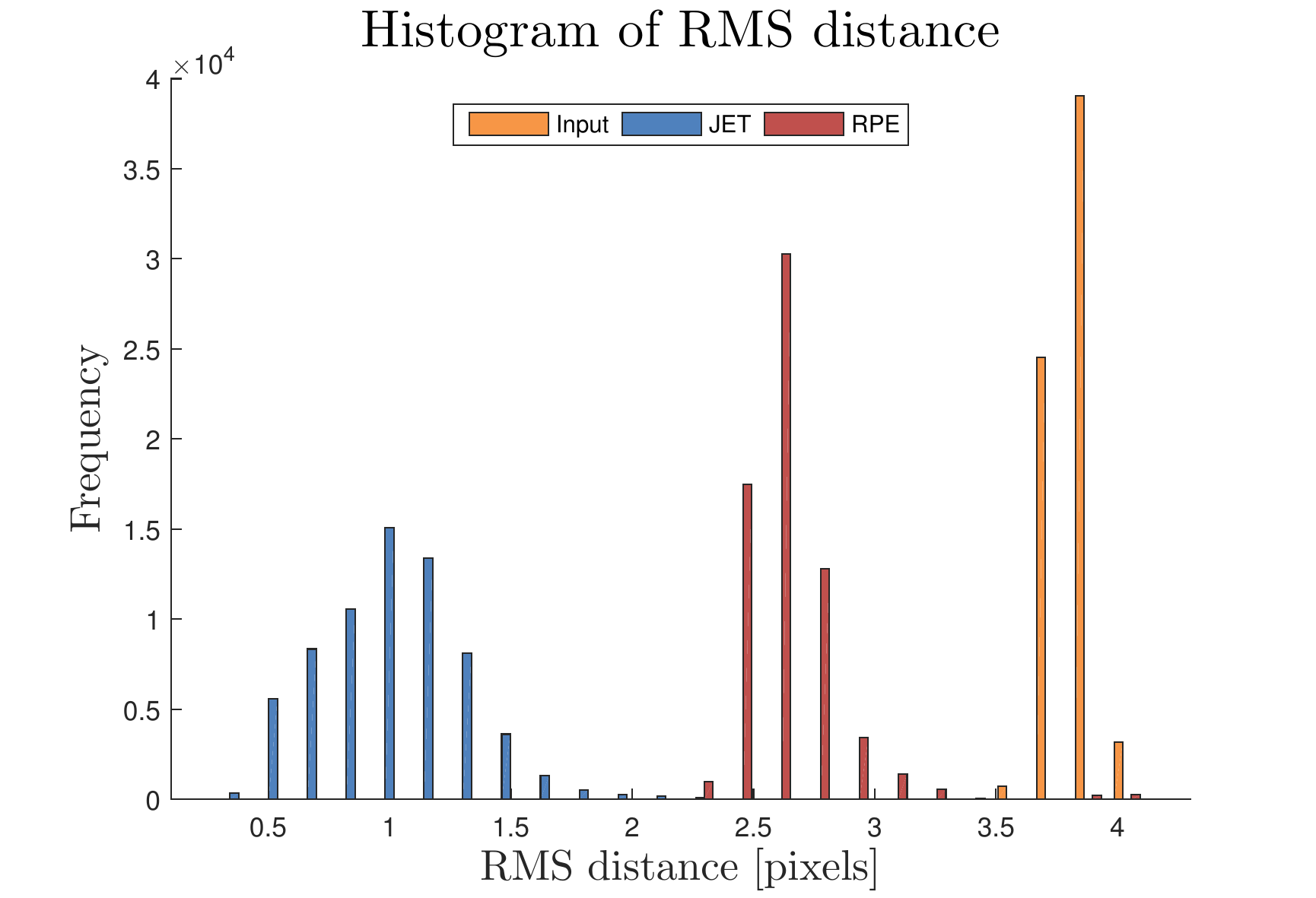}
	\includegraphics[width=.33\linewidth]{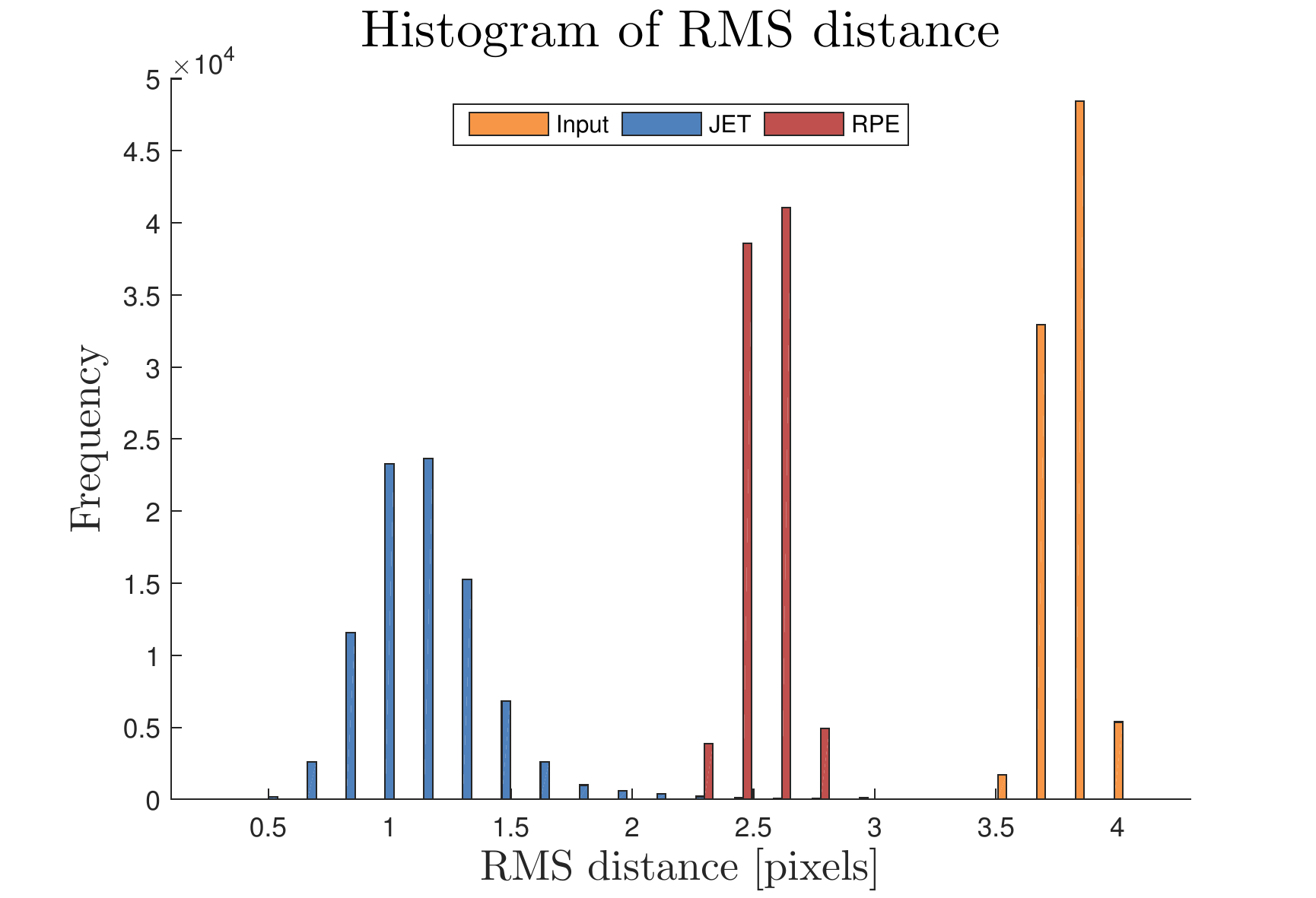}
	\includegraphics[width=.33\linewidth]{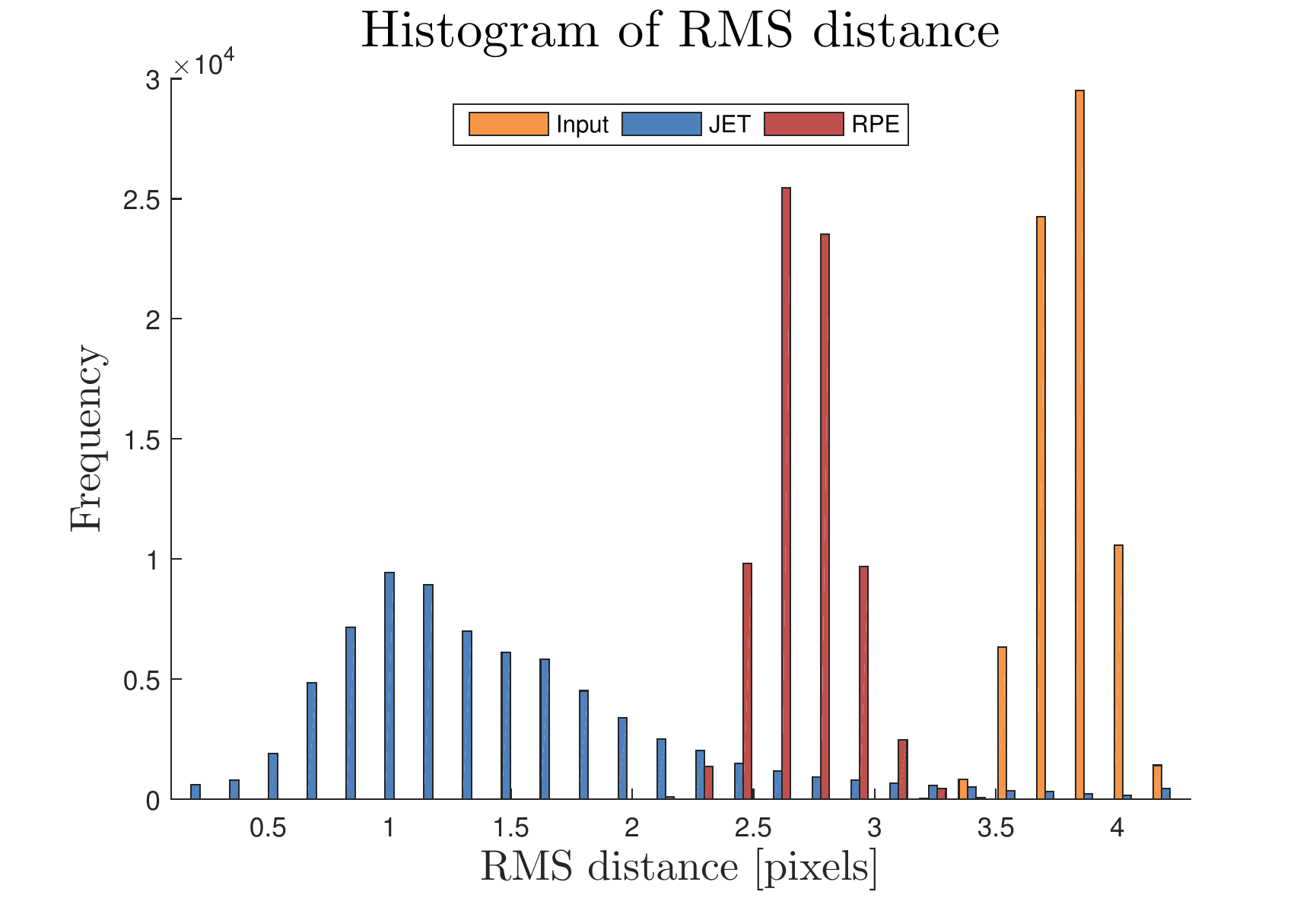}
	\caption{Evaluation on the COnGRATS sequence `ConstructionSite` using prior knowledge ($\xi_{Q} = 1$, $\xi_{R} = 0.5$) (first column) and evaluation on the COnGRATS sequence `Highway` and on the RGB-D sequence `LivingRoom02` without using prior knowledge (second and third column). First and second row visualize the distributions of the quality measures $\rho$ (rotational error) and $\Omega$ (translational error) of the relative pose. The third row exhibits the SSD measure (photometric error) over the courses of the sequences and the last row visualizes the distribution of the RMS measures (positional error) of the correspondences.}
	\label{fig:evaluation}
\end{figure*}

The COnGRATS dataset contains two road scenes of a construction site on a highway (`ConstructionSite`) showing maneuvers at low velocities and another highway scene (`Highway`) with the car travelling mainly straight ahead at a much higher speed. Both scenes use a setup of the camera similar to KITTI \cite{GeigerLenzStillerUrtasun2013IJRR} and were generated using the pose information from the KITTI odometry dataset \cite{GeigerLenzUrtasun2012CVPR}. This enables us to use the extensive motion data in KITTI to generate a statistical model of ego-dynamics to be used as statistical prior.
The results are shown in the first and second column of figure \ref{fig:evaluation} and the mean and standard deviation are listed in table \ref{tab:mean-and-standard-deviation-rho-omega}.
The results show that JET, using image information, reduces the rotational error $\rho$ to approximately the half of the value of RPE without using prior knowledge. While using prior knowledge does not seem to have a large impact on the optimization of the rotation of RPE, it does have it for JET. Using the prior, JET is able to nearly halve the rotational error once more, compared to not using a prior. The observations for JET are also true for the translational error $\Omega$: the use of a prior more than halves the error. In contrast to the optimization of the rotation, the translation optimization of RPE also greatly benefits from using the prior, leading to a reduction of the error by more than a half. This behavior becomes very clear when comparing the histograms of $\rho$ and $\Omega$ for the cases with and without prior information (first and second column of figure \ref{fig:evaluation} respectively). JET is the clear winner for the rotation optimization and also dominates the optimization of the translation without using the prior. Enabling the prior leads to a head to head situation for the translational error.

Regarding the SSD, it is very easy to see the influence of the optional Lucas-Kanade tracking for RPE. The value is strongly decreased. However, JET also dominates this area. It achieves SSD values that are clearly below the ground truth value indicating a very good quality of the optimization of the feature correspondences. Nevertheless, on an average the feature correspondences of JET deviate by about 1 pixel off the ground truth position. The reason for this behavior (similar and even better SSD value while still deviating from the ground truth position) can be explained by the use of patch matching and the existence of a locally non-constant optical flow field (caused by rotation and translation in the direction of the optical axis leading to different scalings). Apart from that, the RMS value of JET is clearly superior to the results of RPE.

The ICL-NUIM RGB-D dataset we evaluated on contains synthetic data of a hand held camera which is carried through a living room (`LivingRoom02`) and an office room (`OfficeRoom02`). The motion is dominated by strong rotations and involves only slight translation. As the motion is less constrained, compared to vehicle motions, the positive influence of integrating prior knowledge is less pronounced. Therefore, we only present results without using the prior ($\xi_{q} = 0$, $\xi_{R} = 0$). They are visualized in the third column of figure \ref{fig:evaluation} and listed in table \ref{tab:mean-and-standard-deviation-rho-omega}.

In summary, the results of the RGB-D dataset are similar to the ones achieved on COnGRATS. JET is superior to RPE in optimizing the rotation and translation (see histograms in third column of \ref{fig:evaluation}). It is dominating the SSD results by achieving SSD values below the ground truth value and it is also clearly superior in optimizing the image point correspondences (RMS). Due to the harder requirements 
of data from a hand held camera, all results are slightly worse than they were for the COnGRATS dataset. Especially the optimization of the translation direction is very tough (see $\Omega$ in third column of figure \ref{fig:evaluation} and table \ref{tab:mean-and-standard-deviation-rho-omega}), when only slight magnitudes of the translation can be observed. Already a minor shaking of the hand, as it is simulated in the scenes, can lead to constantly and much pronounced changes in the direction of the translation. Even though the effect of this behavior only has a small influence on the relative pose and the optical flow in the image domain, it has a strong influence when looking at the evaluation of the direction of the translation. This is a limitation of our parametrization: the direction of the translation is almost undetermined due to its vanishing magnitude, and no scale is available due to the use of a mono camera setup.

Apart from this, the optimization of the unscaled relative pose and the feature correspondences was very successful and largely improved by including the photometric matching information when using JET.

\subsection{KITTI Dataset}

We also performed experiments on the KITTI dataset. Since KITTI does not provide ground truth for image point correspondences (\eg via a dense depth or optical flow map), we cannot use ground truth for the correspondences and apply noise to them to serve as an initializiation. Therefore, we initialize the correspondences by employing propagation based tracking as presented in \cite{FananiOchsBradlerMester2016IV}. Similar to the experiments on the synthetic data, we compare the results of both methods. We use the KITTI ground truth of the pose and compare JET and RPE with respect to the rotational error $\rho$ and the translational error $\Omega$. In order to compare the quality of the feature correspondences of the two methods, we regard the photometric error (SSD).

The results of the experiment on the KITTI dataset are shown in table \ref{tab:result-kitti}. The table presents the mean values of the Rodrigues angle ($\rho$), the angle of intersection of the translation ($\Omega$), and the SSD for each KITTI sequence that has ground truth available. The results confirm that JET performs clearly better than RPE in matters of rotation optimization. The mean of the rotational error $\rho$ is two to three times lower than the one of RPE. In terms of translation, the results show a head and head situation of RPE and JET with a slight lead of JET. Thus, in summary JET yields a significantly better pose than RPE.

Besides improving the pose, JET also refines the feature correspondences. However, as correspondence ground truth is not available in KITTI, the residual error in feature correspondences after performing JET cannot be determined. However, the feature correspondences from JET possess a much smaller photometric error (SSD) than after RPE optimization as can be seen in table \ref{tab:result-kitti}.
\section{Summary and Conclusion}

This paper proposed a novel algorithm in the area of feature tracking and frame-to-frame pose estimation,
 denoted as Joint Epipolar Tracking (JET).
The proposed algorithm employs a direct method to simultaneously optimize the epipolar geometry
and feature correspondences.
It iteratively solves the minimization problem of the newly introduced joint loss function
 where additional statistical information about the motion can be included to serve as prior knowledge.
 The proposed method has been shown to perform better
 than the competing method of RPE optimization
 by experiments on several datasets,
synthetic and real, such as COnGRATS, ICL-NUIM, and KITTI.
It attains real-time performance:
approximately $30$fps utilizing roughly
400 features with patch size of $9 \times 9$
 pixels
 on a single
  thread
 of an Intel Core i7-6700 CPU.
On an average, the rotational errors are three times smaller compared to RPE.
The translation direction can be improved as well if the translation is sufficiently encoded in the optical flow of the image.
Furthermore, the photometric error (SSD) of the feature patches
is massively reduced in all cases which suggest a better quality
also of the 3D information that can be computed from the point correspondences.



\end{document}